\documentclass[sigconf]{aamas} 

\usepackage{balance} 

\usepackage{algorithm}
\usepackage{algorithmic}
\usepackage{makecell}
\usepackage{url}
\usepackage{subfig}

\setcopyright{ifaamas}
\acmConference[AAMAS '23]{Proc.\@ of the 22nd International Conference
on Autonomous Agents and Multiagent Systems (AAMAS 2023)}{May 29 -- June 2, 2023}
{London, United Kingdom}{A.~Ricci, W.~Yeoh, N.~Agmon, B.~An (eds.)}
\copyrightyear{2023}
\acmYear{2023}
\acmDOI{}
\acmPrice{}
\acmISBN{}

\acmSubmissionID{758}

\title{Preference-Aware Delivery Planning for Last-Mile Logistics}

\author{Qian Shao}
\affiliation{
  \institution{School of Computing and Information Systems\\
  Singapore Management University}
  \city{Singapore}
  \country{Singapore}}
\email{qianshao.2020@phdcs.smu.edu.sg}

\author{Shih-Fen Cheng}
\affiliation{
  \institution{School of Computing and Information Systems\\
  Singapore Management University}
  \city{Singapore}
  \country{Singapore}}
\email{sfcheng@smu.edu.sg}

\begin{abstract}
Optimizing delivery routes for last-mile logistics service is challenging and has attracted the attention of many researchers. These problems are usually modeled and solved as variants of vehicle routing problems (VRPs) with challenging real-world constraints (e.g., time windows, precedence). However, despite many decades of solid research on solving these VRP instances, we still see significant gaps between optimized routes and the routes that are actually preferred by the practitioners. Most of these gaps are due to the difference between what's being optimized, and what the practitioners actually care about, which is hard to be defined exactly in many instances.
In this paper, we propose a novel hierarchical route optimizer with learnable parameters that combines the strength of both the optimization and machine learning approaches. Our hierarchical router first solves a zone-level Traveling Salesman Problem with learnable weights on various zone-level features; with the zone visit sequence fixed, we then solve the stop-level vehicle routing problem as a Shortest Hamiltonian Path problem. The Bayesian optimization approach is then introduced to allow us to adjust the weights to be assigned to different zone features used in solving the zone-level Traveling Salesman Problem. By using a real-world delivery dataset provided by the Amazon Last Mile Routing Research Challenge, we demonstrate the importance of having both the optimization and the machine learning components. We also demonstrate how we can use route-related features to identify instances that we might have difficulty with. This paves ways to further research on how we can tackle these difficult instances.
\end{abstract}


\newcommand{\BibTeX}{\rm B\kern-.05em{\sc i\kern-.025em b}\kern-.08em\TeX}

\begin{document}


\pagestyle{fancy}
\fancyhead{}


\maketitle 

\section{Introduction}
Optimizing delivery routes for last-mile logistics service is challenging and has attracted the attention of many researchers. Most existing approaches from the literature (e.g., see \cite{caceres2014rich}) aim to solve a wide variety of vehicle routing problems (VRPs), usually with many complicating real-world constraints (e.g., time windows, precedence). However, despite many decades of solid research on how to solve these VRP instances as efficiently and effectively as possible, we still see significant gaps between optimized routes and the routes that are actually preferred by the practitioners.

Most of these gaps are due to the difference between what's being optimized, and what the practitioners actually care about. From the literature, typical objectives being optimized could include travel time, distance, or cost. However, experienced delivery drivers have first-hand knowledge of the area and the customers that they are serving, and thus could plan their routes based on a wide variety of additional factors that are difficult to formalize and quantify. As a result, drivers often deviate from the planned routes. While drivers could alter the actual routes to satisfy their own constraints or incorporate their personal knowledge, they could potentially sacrifice overall system metrics, such as fuel consumption, delivery time, and the order packing operations (onto the delivery vehicle) that are closely related to the routing sequence. Therefore, it will be much more desirable if we could incorporate drivers' tacit knowledge about the delivery area into our route planning algorithm.

In this paper, we propose a novel hierarchical router with learnable parameters that combines the strength of both the optimization and machine learning approaches. Our hierarchical router first solves a zone-level Traveling Salesman Problem (TSP) with learnable weights on various zone-level features; with the zone visit sequence fixed, we then solve the stop-level vehicle routing problem as a Shortest Hamiltonian Path problem. The Bayesian optimization approach is then introduced to allow us to adjust the weights to be assigned to different zone features used in solving the zone-level TSP. By using a real-world delivery dataset provided by the Amazon Last Mile Routing Research Challenge \cite{merchan_2021_2022}, we demonstrate the importance of having both the optimization and the machine learning components. A critical difference between the problem we are solving and most past work is how we evaluate the solution quality, which is based on how close our generated route sequences are to the highly-rated route sequences from the historical dataset. 
Our key contributions are summarized as follows:
\begin{itemize}
    \item We proposed a novel hierarchical route optimizer with learnable parameters that combines the strength of both the optimization and machine learning approaches.\footnote{The implementation can be found at \url{https://github.com/SHAOQIAN12/HR-LP.git}.}
    
    \item We demonstrate the effectiveness of our approach using a real-world delivery dataset provided by the Amazon Last Mile Routing Research Challenge.
    
    \item Finally, we also demonstrate how we can use route-related features to identify instances that we might have difficulty with.
\end{itemize}

\section{Related Work}
\subsection{Last-mile Delivery Problem} The challenging target service levels, the small dimension of parcels, and the high level of dispersal of destinations make the last-mile delivery problem become a tricky part of the delivery process~\cite{macioszek2017first}. Therefore, how to increase the efficiency of last-mile delivery has received growing attention in recent years. Some research focuses on how to optimize the traditional delivery mode and propose different versions of the vehicle routing problem that calculate the optimal route to deliver a set of demands to dispersed destinations~\cite{agussurja_state_2019}. Some methods focus on innovative solutions to increase the efficiency of last-mile delivery, such methods include parcel lockers~\cite{iwan2016analysis}, crowdsourcing logistics~\cite{cheng2017scalable,han2021exact}, and drone-based solutions~\cite{salama2020joint}. 

\subsection{Traveling Salesman Problem} The problem we study in this paper is similar to the classical TSP, which aims to find the shortest tour that visits all cities in a given set exactly once and returns to the origin. The TSP has been extensively studied~\cite{gutin2006traveling} and has been proved as an NP-hard problem~\cite{hartmanis1982computers}. Classic approaches to solving the TSP can be classified into exact methods and heuristic methods. The former has been studied using integer linear programming (ILP)~\cite{10.2307/j.ctt7s8xg} that are guaranteed to find an optimal solution, but ILP problems are computationally expensive to be used in practice. To address this computational challenge, many heuristic methods are proposed, which include the Savings algorithm~\cite{clarke1964scheduling}, Tabu Search \cite{glover1986future}, Greedy Randomized Adaptive Search Procedure \cite{feo1989probabilistic}, Simulated Annealing \cite{vcerny1985thermodynamical,kirkpatrick1983optimization}, and Genetic Algorithms~\cite{holland1992adaptation}.

More recently, deep learning approaches have been proposed as part of the heuristic for solving hard combinatorial optimization problems such as the TSP \cite{BENGIO2021405}. For example, Pointer Networks (PtrNet) \cite{vinyals2015pointer} learns a sequence model coupled with an attention mechanism trained to output TSP tours using solutions generated by Concorde \cite{10.2307/j.ctt7s8xg}. The PtrNet is further extended to learn without supervision using Policy Gradient, trained to output a distribution over node permutations~\cite{https://doi.org/10.48550/arxiv.1611.09940}. 

As we aim to generate route sequences that are similar to highly rated routes in practice, our objective is different from that of the classical TSP instances and thus past approaches are not directly applicable.

\subsection{Bayesian Optimization} Bayesian optimization is widely adopted for hyper-parameter tuning in optimizing objective functions. It is applied widely in areas such as combinatorial optimization~\cite{hutter2011sequential,wang2013bayesian}, automatic machine learning  \cite{ NIPS2011_86e8f7ab,snoek2012practical,thornton2013auto, 10.5555/2999792.2999836} and reinforcement learning  \cite{https://doi.org/10.48550/arxiv.1012.2599}. Bayesian optimization assumes the unknown function is sampled from a Gaussian process and maintains a posterior distribution for this function as observations are based~\cite{snoek2012practical}. Then an acquisition function is used to decide where to sample for the next step. The acquisition function measures the value that would be generated by the evaluation of the objective function at a new point, based on the current posterior distribution. The acquisition function can be \textit{expected improvement}, \textit{entropy search}, and \textit{knowledge gradient}. In our work, we use \textit{expected improvement} as the acquisition function.

\section{Problem Description}
Our problem formulation is inspired by the dataset provided by the Amazon Last Mile Routing Research Challenge \cite{merchan_2021_2022}. The objective is to learn from experienced delivery drivers so that the last-mile route optimizer is capable of generating optimized routes that drivers would also rate highly. To concretely and quantitatively define what constitutes a good route (from the driver's perspective), Amazon provides a labeled dataset where around 6,000 actual delivery routes are rated by drivers to be of high-, medium-, and low-quality (more details on the dataset will be presented shortly). The evaluation of generated routes depends on how closely a generated route resembles a highly rated benchmark route from history.

\subsection{Dataset Description}
The Amazon dataset\footnote{The Amazon dataset is available at \url{https://registry.opendata.aws/amazon-last-mile-challenges/}} contains historical delivery routes from five metropolitan areas in the USA: Austin, Boston, Chicago, Los Angeles, and Seattle. In the dataset, there are 2718, 3292, and 102 routes that are of high-, medium-, and low-quality respectively. The dataset contains three major components: 1) the routes, 2) the stops, and 3) the packages. We elaborate on important information in all three components below.

\subsubsection{The Route Information}
For each route, we have: a) a unique route ID, b) the departure time, c) the volume capacity of the vehicle, d) the actual sequence of stops visited, and e) the rating label, which can be high, medium, or low. 

\subsubsection{The Stop Information}
For each stop, we have: a) a unique stop ID, b) the geographical coordinate of the stop, and c) the zone ID (a planning unit defined by the dataset) where this stop is located. The travel times between stops are also provided.

\subsubsection{The Package Information}
For each package, we have: a) a unique package ID, b) the status, which can be rejected, delivered, or delivery-attempted, c) the stop ID this package belongs to, d) the dimensions (maximal width, length, and height), e) delivery time window (if any), and f) the estimated service time. 

\subsection{Evaluation Metric}

The evaluation metric is whether a generated route closely resembles the highly-rated benchmark route. 
Formally speaking, the \emph{score} of a route is composed of two components: the \emph{Sequence Deviation} (SD) and the \emph{Edit Distance with Real Penalty} (ERP). The SD component measures how different a given sequence is from the benchmark sequence, and takes values from 0 to 1, with 0 indicating that the two sequences are identical. On the other hand, the ERP component measures how many single-element operations (insertions, deletions, and substitutions) are required to transform the given sequence into the benchmark sequence. To reflect the travel distance, the operations are further weighted by the physical distance among affected stops. In other words, the ERP component measures how far apart a given sequence is from the benchmark sequence.

Let $A$ be the benchmark sequence, and $B$ be the sequence we would like to evaluate, the \emph{route score} of $B$ is defined as:
\begin{equation}
    \label{score calculation}
    route\_score(A,B) = \frac{SD(A,B) \cdot {ERP}_{n}(A,B)}{{ERP}_{e}(A,B)}.
\end{equation}
The $SD$ component is defined as:
\begin{equation}
    \label{SD}
    SD(A,B) = \frac{2}{n(n-1)} \sum_{i=1}^n|a_i - a_{i-1}| - 1,
\end{equation}
where $n$ is the total number of stops in the sequence ($A$ and $B$ should have equal number of stops), and $a_i$ is the position where the $i^{\text{th}}$ stop in $B$ appears in $A$.

The ${ERP}_{n}(A,B)$ component captures normalized travel time between stops at the same location of both sequences and is recursively defined as:
\begin{equation}
    \label{ERP}
    {ERP}_{n}(tail(A),tail(B))+ {time}_{n}(A[0],B[0]),
\end{equation}
where ${time}_{n}(x,y)$ is the normalized travel time between stop $x$ and stop $y$, $tail(X)$ is the sequence $X$ without its first stop, $X[0]$ is the first stop in the sequence $X$. The implicit assumption of the above definition is that $A$ and $B$ should include the same set of stops.

The ${ERP}_e(A,B)$ component captures the number of edit operations (insertions, substitutions, or deletions) required to transform sequence $B$ to sequence $A$. The ratio ${ERP}_n(A,B) / {ERP}_e(A,B)$ can be intuitively interpreted as the average normalized travel time per edit operation.

The performance of the routing engine is simply the average score of all routes:
\begin{equation}
    \label{submission score calculation}
    score \!=\! \frac{1}{|I|}\sum_{i \in I} route\_score_i,
\end{equation}
where $I$ contains all route instances being evaluated.

\section{Solution Approach}

\begin{figure*}[!htb]
    \centering
    \subfloat[Benchmark routing sequence.\label{fig:benchmark}]{%
        \includegraphics[height=2.65in]{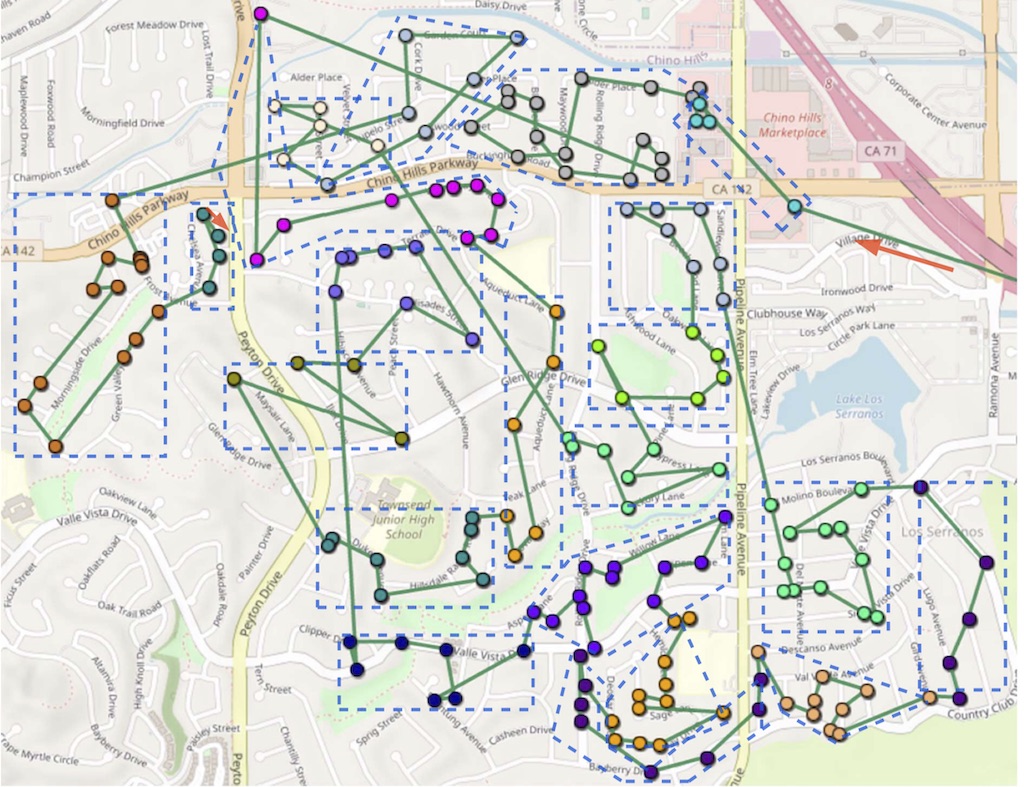}}\;\;
    \subfloat[Routed by a TSP heuristics.\label{fig:TSP}]{%
        \includegraphics[height=2.65in]{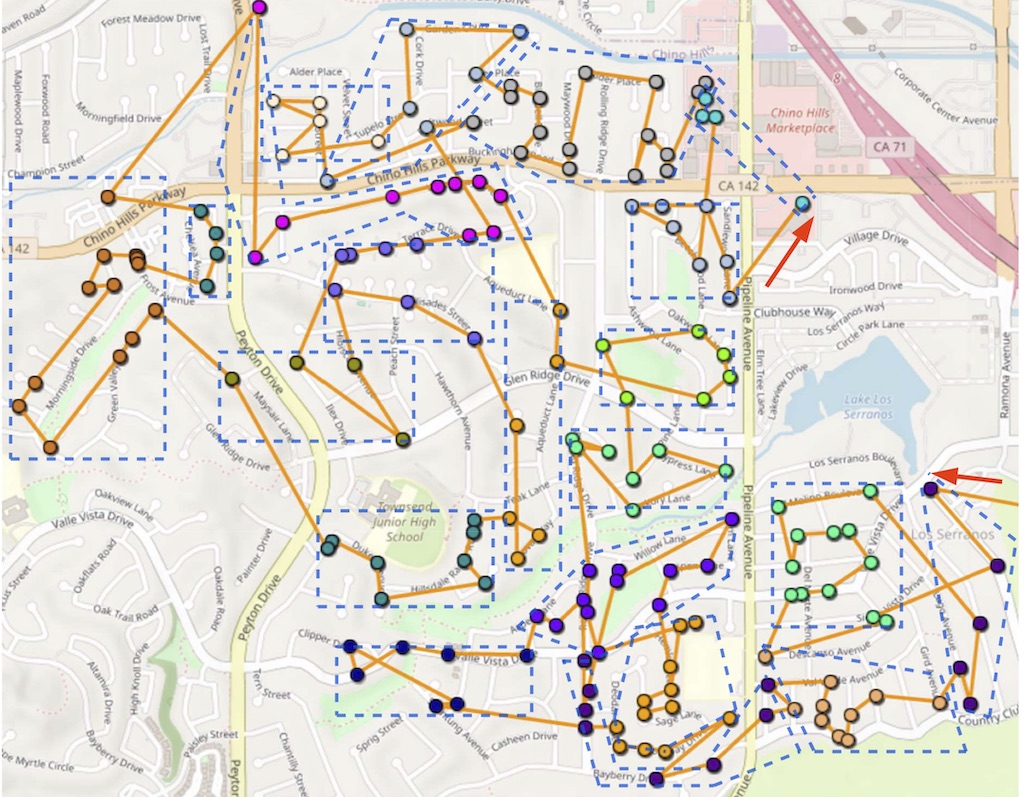}}
    \caption{How a typical benchmark sequence differs from the TSP route.}
    \label{fig:zone_base}
\end{figure*}
To explore potential design ideas that would optimize the routing of stops and at the same time incorporate drivers' tacit knowledge of what constitutes a good route, we sample a wide collection of routes that are highly rated by drivers and compare them against routes generated via a simple TSP heuristic (see Figure~\ref{fig:zone_base} for a typical example). What we see from the comparison is strong visual evidence that drivers prefer to schedule their visits in blocks of zones, i.e., visiting all stops within a zone before moving on to the next zone. While the routes generated by the TSP heuristic tend to mix stops from different zones. 

Given this observation, we propose to design a hierarchical router, where the zone visitation sequence is decided first, after which stops within respective zones are then routed. The most important information we would harvest from benchmark routes is how drivers would sequence zones. To achieve this, we identify a collection of features associated with zones and define the cost function at the zone level as a linear combination of zone features (with feature weights as parameters). We use the Savings algorithm to find out the zone visitation sequence, and utilize a Bayesian optimization method to iteratively update the weights associated with all zone features using historical benchmark sequences.

\subsection{A Hierarchical Router with Learnable Parameter}

We call our approach the Hierarchical Router with Learnable Parameters (HR-LP), since it generates a route at the zone level first, before progressing to route the stops within each zone. To formulate the zone sequencing problem, we define $c_{ij}$ to represent the inclination a driver would have in choosing zone $j$ when in zone $i$ (the lower the $c_{ij}$, the higher the inclination). We define $c_{ij}$ as:
\begin{equation}
    \label{cost features function}
    c_{ij} = \sum_{k=1}^{K} \theta_k \varphi_k(i,j),
\end{equation}
where $\varphi_k(i,j)$ and $\theta_k$ represent the value of feature $k$ and the weight associated with feature $k$ respectively. The feature weights will be learned via a Bayesian optimization approach, and we will defer its introduction to the later section.

We have in total 14 zone features, which can be either native (purely based on the characteristics of the zone) or derived. Among these 14 feature candidates, we select 5 features that are most significant, which are introduced below: 
\begin{itemize}
    \item $\varphi_1(i,j)$: average travel time from any stop in zone $i$ to any stop in zone $j$.

    \item $\varphi_2(i,j)$: Euclidean travel distance from the centroid of zone $i$ to the centroid of zone $j$. The x-coordinate of the zone centroid is calculated as: $x_c = \frac{\sum_{i=1}^n x_i}{n}$, in which $n$ is the number of stops in the zone, and $x_i$ is the x-coordinate of stop $i$. The y-coordinate of the centroid is calculated similarly.

    \item $\varphi_3(i,j)$: the ratio of the average travel time from the depot to all stops in zone $j$ over the average travel time from the depot to all stops in zone $i$.

    \item $\varphi_4(i,j)$: the ratio of the average travel time from all stops in zone $j$ to depot over the average travel time from all stops in zone $i$ to depot.

    \item $\varphi_5(i,j)$: 1 if zones $i$ and $j$ are in the same main zones, and 0 otherwise. (The zone ID is denoted as {X-N.MY}, and two zones are considered to be in the same main zone if their X and N are identical.)
\end{itemize}
With $c_{ij}$ defined, we can formulate the zone sequencing problem following the standard TSP formulation:
\begin{align}
\min & \sum_{i\in N} \sum_{j\in N} c_{ij} z_{ij}, \label{cost function}\\
\text{s.t.}\;\;\sum_{j\in N} z_{ij} & = \sum_{j\in N} z_{ji} = 1   , \forall i \in N, \label{flow balance}\\
z_{ii} & = 0 , \forall i \in N, \label{self loop elimination}\\
\sum_{i\in R} \sum_{j\in R} z_{ij} & \leq \left|R\right|-1 , \forall R \subseteq N, \label{subtour elimination}\\
z_{ij} & = \{0,1\}  , \forall (i,j) \in A, \label{decision variable}
\end{align}
where the decision variable $z_{ij} = 1$ represents that the driver traverses from zones $i$ to $j$, \eqref{flow balance} ensures that each zone will be visited exactly once, \eqref{self loop elimination} eliminates self loop, and \eqref{subtour elimination} prevents subtours.

We use the Savings algorithm \cite{clarke1964scheduling} to solve the above zone-level TSP and the open-source OR-tools solver \cite{ortools} is used for the implementation. The optimal zone visit sequence is denoted as $Z^* = (0,z_1,z_2,...,z_n)$, where $0$ represents the depot. With the zone visit sequence, we then solve for the stop visit sequence within each zone and decide which stops should be used as the starting and ending stops (to connect to the next and from the previous zones). 

Since the starting and the ending stops are different when routing stops within a zone, the problem is formulated as a Shortest Hamiltonian Path (SHP) problem below:
\begin{align}
\label{stop cost function}
\min & \sum_{i\in M} \sum_{j\in M} t_{ij} x_{ij},\\
\label{stop flow balance}
\text{s.t.} \;\; \sum_{j\in M} x_{ij} & = \sum_{j\in M} x_{ji} = 1 , \forall i \in M \backslash \{m_o,m_d\},\\
\label{start end stop balance}
\sum_{i\in M} x_{im_d} & = \sum_{j\in M} x_{m_oj} = 1,\\
\label{start end stop balance 1}
\sum_{i\in M} x_{im_o} & = \sum_{j\in M} x_{m_dj} = 0,\\
\label{stop subtour elimination}
\sum_{i\in R}\sum_{j\in R} x_{ij} & \leq \left|R\right|-1, \forall R \subseteq M,\\
\label{stop decision variable}
x_{ii} = 0 , \forall i & \in M, \;\; x_{ij} = \{0,1\} , \forall (i,j) \in A.
\end{align}
The stop-level routing problem for each zone is similar to the zone-level TSP formulation. For each zone, given the directed graph $G = (M,A)$, where $M$ denotes the location of $m$ stops in the zone. The decision variable $x_{ij} = 1$ represents that the driver traverses from stops $i$ to $j$, with $t_{ij}$ representing the travel time. The objective function ~\eqref{stop cost function} minimizes the total travel time within the zone. We denote $m_o$ and $m_d$ as the first stop when entering the zone and the last stop when leaving the zone; this is enforced by ~\eqref{start end stop balance 1}. 

We further elaborate on the process of generating the complete stop sequence below:
\begin{itemize}
    \item \textbf{Step 1}: For a route, obtain the optimal zone visit sequence $Z^* = (0,z_1,...,z_n)$ by solving the zone-level TSP. 
    
    \item \textbf{Step 2}: Initialize the optimal stop sequence $S^*$ with the depot `0' as the first element.
    
    \item \textbf{Step 3}: For each stop in zone $z_{i-1}$, calculate the average travel time to all stops $x \in z_i$. Sort stops in $z_{i-1}$ in the ascending order following the computed average travel time.
    Insert the top-$h$ stops into the candidate starting set $A$.
    
    \item \textbf{Step 4}: For each stop in zone $z_{i+1}$, calculate the average travel time from all stops $x \in z_i$. Sort stops in $z_{i+1}$ in the ascending order following the computed average travel time. Insert the top-$h$ stops into the candidate ending set $B$. If zone $z_i$ is the last visited zone, insert depot `0' into $B$.

    \item \textbf{Step 5}: For each pair of possible starting-ending stops from $A \times B$, solve the corresponding stop-level SHP. Except for the first and the last zones, we should have $h^2$ stop-level SHP instances for each zone (each instance comes with different starting and ending stops).
    
    \item \textbf{Step 6}: Choose the stop sequence with minimum total travel time as the optimal sequence $S^*_i$ for zone $i$.
    
    \item \textbf{Step 7}: For each zone $z_i$ in $Z^*$, repeat \textbf{Steps 3-6} to obtain the optimal stop sequence $S^*_i$ and add it into the optimal stop sequence $S^*$.
\end{itemize}
As before, we utilize the Savings algorithm \cite{clarke1964scheduling} to solve all stop-level SHP instances and OR-tools solver \cite{ortools} is used for the implementation. $h$ is a hyper-parameter which will be discussed in the following experiments.

\begin{algorithm}[!htb]
\caption{Bayesian Optimization}
\label{Algorithm Bayesian Optimization}
\textbf{Input}: Number of initial points $n_0$, route data $D_M$, number of routes $M$, iteration $N$, Gaussian process prior $p(l)$\\
\textbf{Output}: Optimal weight $\boldsymbol\theta^*$ 
\begin{algorithmic}[1] 
\STATE  Compute $n_0$ random initial points: \\ 
$\Psi =\{ (\boldsymbol\theta_1,-l_1(\boldsymbol\theta_1)),\ldots, (\boldsymbol\theta_{n_0},-l_{n_0}(\boldsymbol\theta_{n_0}))\}$.
\STATE $n \gets n_0$.
\WHILE{$n < N$}
    \STATE Update posterior distribution $P(l|\Psi)$.
    \STATE Select new $\boldsymbol\theta_{n+1}$ by optimizing the acquisition function $\alpha$: $\boldsymbol\theta_{n+1} \! = \! \arg\max\limits_{\boldsymbol\theta} \alpha(\boldsymbol\theta,\Psi)$.
    \FOR{$m \in M$}
        \STATE Solve the hierarchical TSP to obtain the proposed sequence of route $m$:  $r^*_m = f(D_m,\boldsymbol\theta_{n+1})$.
    \ENDFOR
    \STATE $l_{n+1} (\boldsymbol\theta_{n+1}) = \frac{1}{M}\sum_{m \in M} route\_score(r^*_m,r_m)$.
    \STATE Augment data: $\Psi \gets \{\Psi,(\boldsymbol\theta_{n+1},-l_{n+1}(\boldsymbol\theta_{n+1}) \} $.
    \STATE $n \gets n+1$.
\ENDWHILE
\STATE \textbf{return}  $\boldsymbol\theta^* =  \arg\max\limits_{\boldsymbol\theta} (-l_1(\boldsymbol\theta_1),\ldots,-l_N(\boldsymbol\theta_N))$
\end{algorithmic}
\end{algorithm}

\subsection{Learning Feature Weights}
As mentioned before, the Bayesian optimization method is used to find the optimal weight parameters $\boldsymbol\theta$ of zone features by iteratively updating the weight parameters and generating route sequences. For the given high-quality routes, we first generate the proposed sequence for each route by executing our HR-LP solver as described before and then calculate the 'score' of the proposed route sequence using \eqref{score calculation}. We then apply the Bayesian Optimization procedures to search for a new set of parameter values in order to minimize the average score of all given routes. These two steps are implemented iteratively to find the optimal weight parameters.

\subsubsection{Bayesian Optimization Algorithm} 
Given a set of known high-quality routes and initial zone feature weights $\boldsymbol\theta$, we generate proposed route sequences $R^* = \{r^*_1,\ldots,r^*_M\}$ by executing the HR-LP solver. The average score can be calculated according to \eqref{SD} -- \eqref{submission score calculation}. We then utilize the Bayesian optimization approach (see Algorithm~\ref{Algorithm Bayesian Optimization}) to update the feature weights iteratively to minimize the average score. In other words, we are minimizing the following objective function:
\begin{equation}
    \label{Bayesian Optimization Objective}
    l(\boldsymbol\theta) = \frac{1}{M} \sum_{m \in M} route\_score(r_m, f(D_m,\boldsymbol\theta)),
\end{equation}
where $M$ denotes the number of routes, $r_m$ is the given benchmark sequence of route $m$, and $f(D_m,\boldsymbol\theta)$ denotes the generated sequence of route $m$ given route data $D_m$ and weight $\boldsymbol\theta = [\theta_1, \ldots, \theta_k]$. Bayesian optimization is thus defined as 
\begin{equation}
    \label{Bayesian Optimization theta}
    \boldsymbol\theta^* = \arg\min_{\boldsymbol\theta \in \boldsymbol\Theta} l(\boldsymbol\theta),
\end{equation}
where $\boldsymbol\Theta$ is all $\theta_k$ such that $1 \leq \theta_k \leq 10$. The actual steps we adopt to solve \eqref{Bayesian Optimization theta} are described in Algorithm \ref{Algorithm Bayesian Optimization}, with \textit{expected improvement} as our acquisition function.

\section{Experiments}

To evaluate our proposed approach, we use the route data from the dataset. We only incorporate 2,718 high-quality routes. There are 17 depots from the dataset, and we train a different set of weight parameters for each of these 17 depots. 
For each depot, we split the route data into 70\% training set and 30\% testing set. In the training phase, we execute the Bayesian optimization process for each depot separately to obtain the optimal weight parameters $\boldsymbol\theta$. In the testing phase, we use the obtained $\boldsymbol\theta^*$ on the unseen testing dataset to generate proposed route sequences by executing our HR-LP solver and obtain the average score. 

To illustrate the importance of having both the optimization and learning components, We compare our proposed approach against the following two baselines:

\textbf{Standard TSP:} We solve the standard TSP using the Savings algorithm directly. For each route, the driver starts from the depot, visits all stops, and returns to the depot. The objective is to minimize the total travel time. The resulting routing sequence is evaluated using the same metric.

\textbf{Stop-level Bayesian Optimization:} In order to illustrate the effectiveness of the hierarchical routing formulation, we use the Bayesian Optimization method to update the weight parameters of stop features, however, we generate the route using only the standard TSP formulation at the stop-level. The TSP formulation is similar to \eqref{cost function}--\eqref{decision variable}. And the stop features we used are designed as follows:
\begin{itemize}
    \item $\varphi_1(i,j)$: travel time from stop $i$ to stop $j$.
    
    \item $\varphi_2(i,j)$: ratio of the travel time from depot to stop $j$ to the travel time from depot to stop $i$.
    
    \item $\varphi_3(i,j)$: ratio of the travel time from stop $j$ to depot to the travel time from stop $i$ to depot.

    \item $\varphi_4(i,j)$: ratio of the number of packages in stop $j$ to the number of packages in stop $i$.
\end{itemize}

\begin{figure}[!htb]
  \centering
  \includegraphics[width=3.35in]{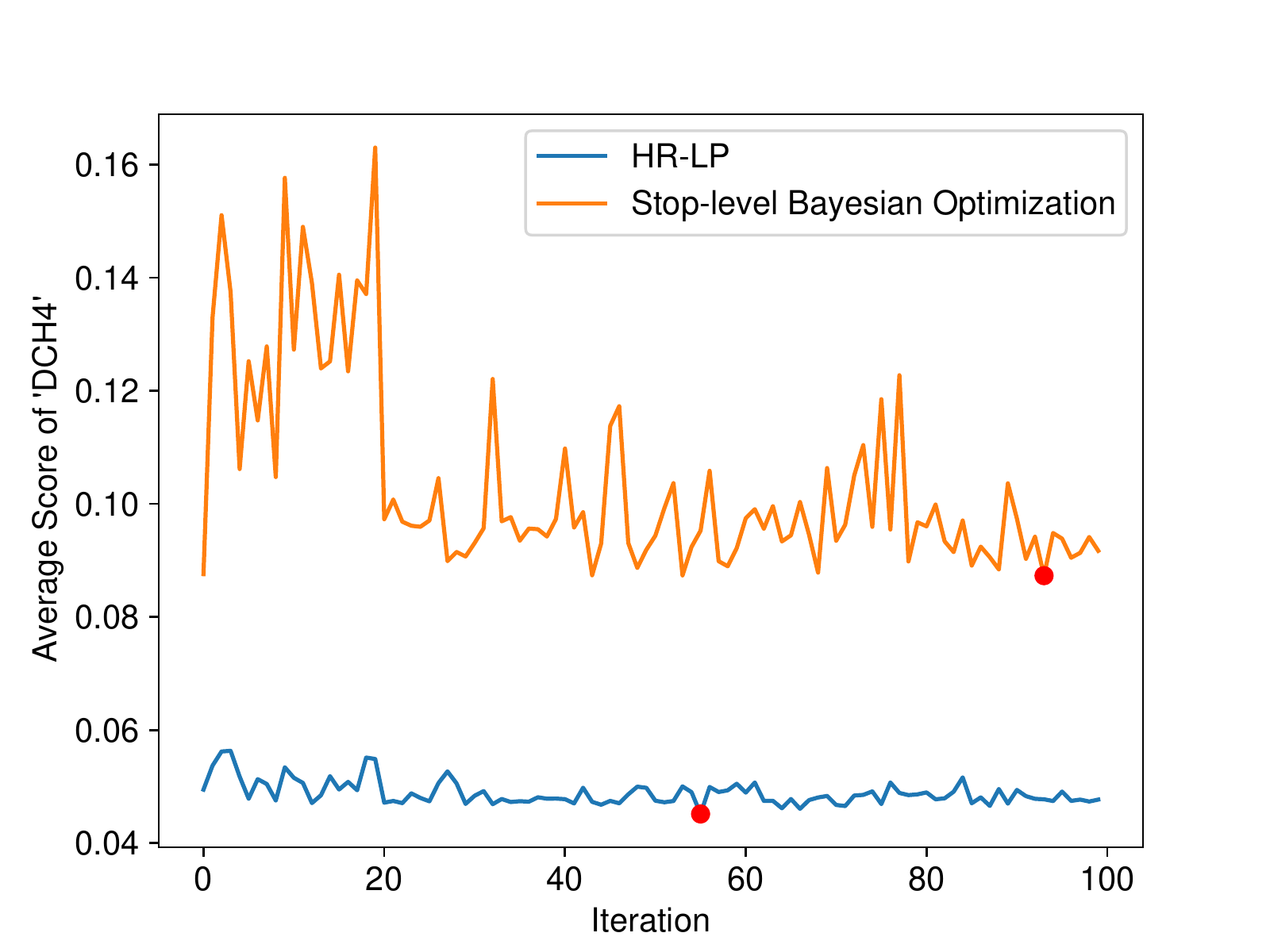}
  \caption{Average score from the Bayesian optimization (for Chicago).}
  \label{tab: iteration} 
\end{figure}
In our experiments, the number of initial points is set to 20, and the iteration number is set to 100. We normalize each feature value into $[0,1]$. Figure \ref{tab: iteration} shows the average score $l(\boldsymbol\theta)$ of training routes departing from depots in Chicago (the results from other cities look very similar). From Figure \ref{tab: iteration} we can see that incorporating our observation that drivers reason at the zone level is crucial in getting good results.

\begin{table}[!htb]
\centering
\caption{Average scores comparison. Lower score means better performance.}
\begin{tabular}{cc}
\hline
        & Testing data              \\
\hline
 \textbf{Standard TSP}       & 0.0898                 \\

 \textbf{Stop-level Bayesian Optimization}  & 0.0992                     \\

  \textbf{HR-LP}   & \textbf{0.0496}     \\
  \hline
\end{tabular}
 \label{tab:Average scores comparison}
\end{table}

\begin{figure*}[!htb]
  \centering
  \includegraphics[width=6.5in]{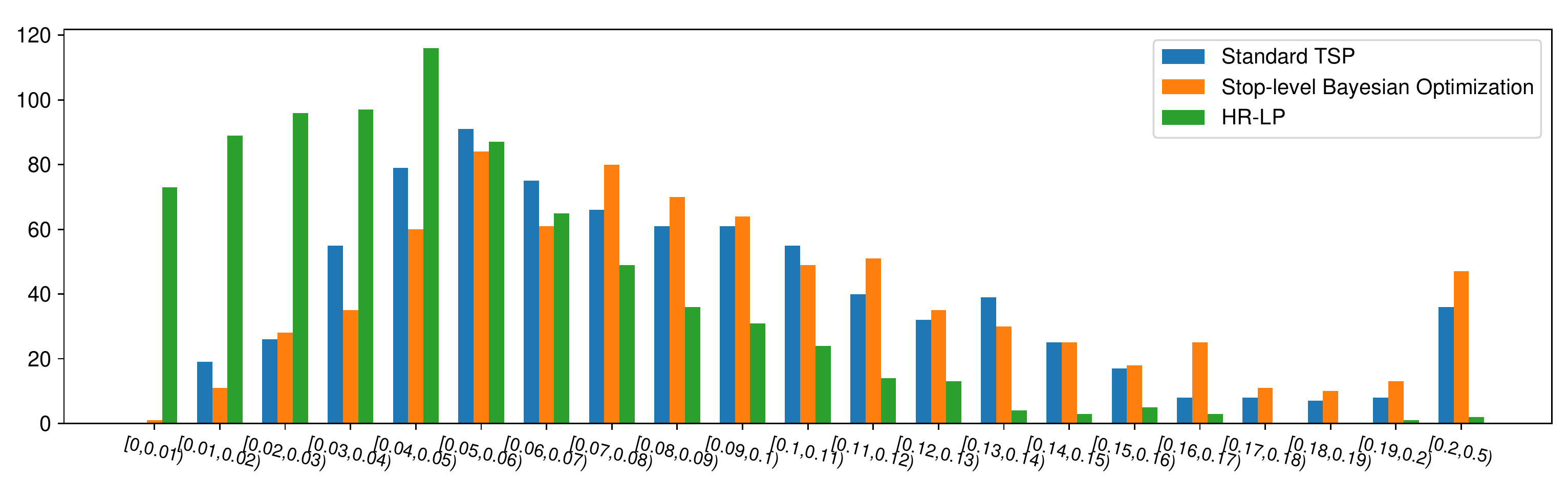}
  \caption{Score distribution for the testing dataset.}
  \label{score distribution} 
\end{figure*}

\begin{figure}[!htb]
\centering
\includegraphics[width=3.35in]{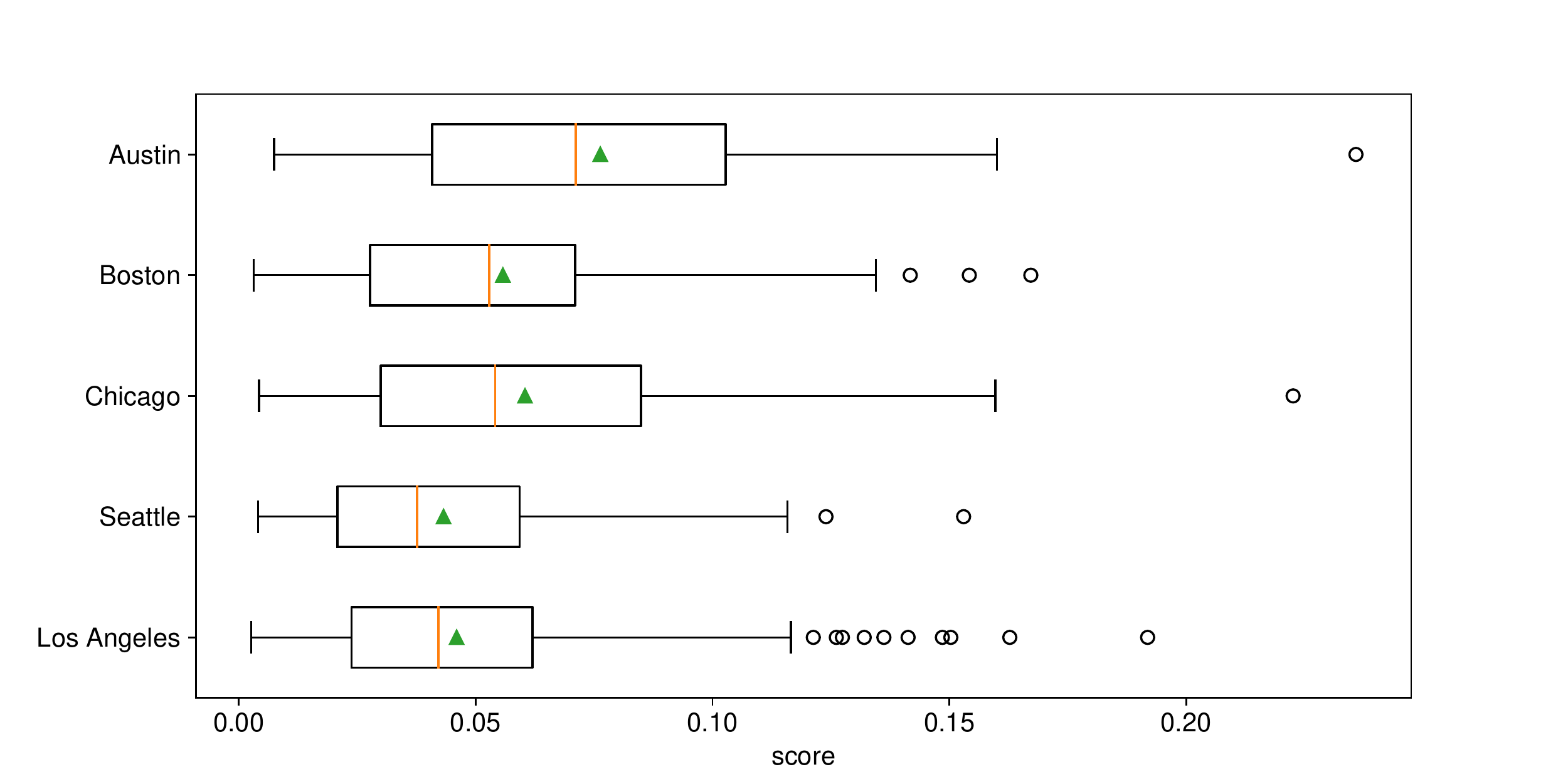} 
\caption{Summary statistics of scores in five cities.}
\label{score by city}
\end{figure}

Figure \ref{score distribution} shows the score distribution of the 808 routes testing dataset, and we can see that our HR-LP approach has the majority of scores lower than 0.06 compared to baselines. Table \ref{tab:Average scores comparison} shows the average score of the testing data using three competing methods. We can see again that our proposed HR-LP approach outperforms the two baselines by nearly 50\%. Figure \ref{score by city} shows the box plot of score distribution in all 5 cities. We notice that the proposed method performs well in Seattle and Los Angeles (with scores lower than 0.05). However, Austin has the highest average score, probably due to insufficient route instances for us to learn the optimal weight parameters. 

We also look at the impact of hyperparameter $h$ (number of candidate starting/ending stops to enumerate) and the computational time. As shown in Table \ref{tab:Average score by setting different k}, increasing the value of $h$ does not seem to impact the outcomes much, which implies that enumerating more starting and ending points do not seem to help us in getting better results. 

\begin{table}[!htb]
\centering
\caption{Mean score and computational time for all $h$. }
\begin{tabular}{cccc}
\hline
   $h$     & Testing data   & Training time & Testing time        \\
\hline
 2      & 0.0496 & 3h12m   & 9m              \\
 3      & 0.0491 &  3h47m  & 9m              \\
 4      & 0.0492 &  4h43m  & 9m              \\
 5      & 0.0489 &  5h47m  & 10m             \\
 \hline
\end{tabular}
\label{tab:Average score by setting different k}
\end{table}

\section{Discussion}

\begin{figure}[!htb]
  \centering
  \includegraphics[width=3.35in]{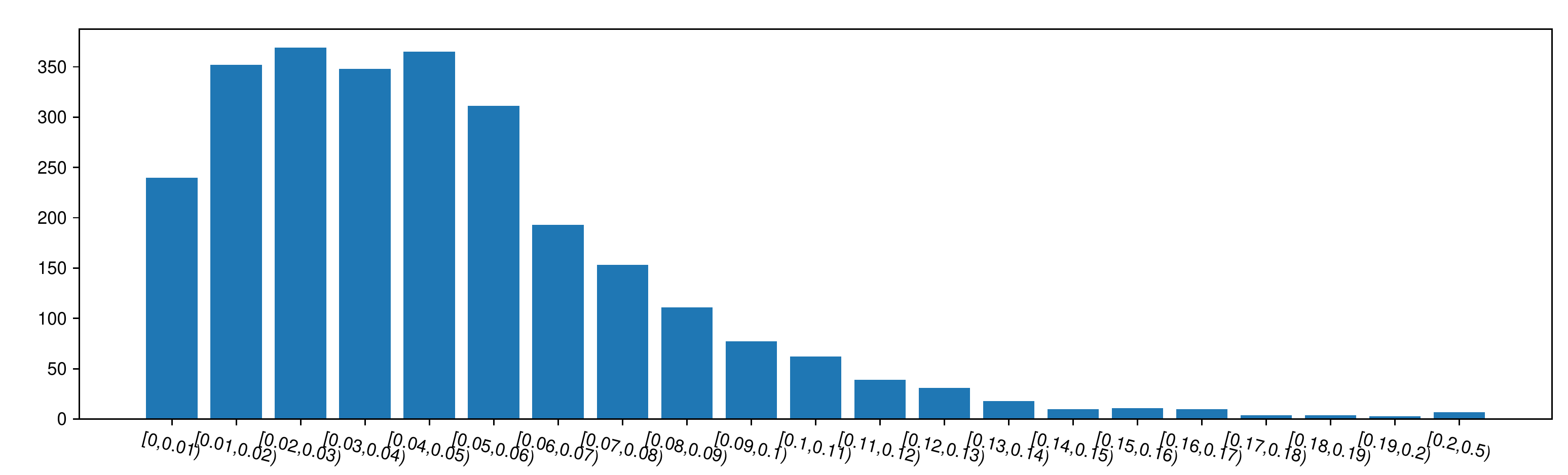}
  \caption{Score distribution of all 2,718 high-quality routes.}
  \label{all score distribution} 
\end{figure}

In this section, we discuss whether the characteristics of route instances would impact the performance of our HR-LP approach. This could help us further improve our proposed approach. We first look at the score distribution of our approach on all 2,718 high-quality routes in Figure \ref{all score distribution}. We can see that while most routes have decent route scores that are below 0.06, there is a long tail in the score distribution. 

To analyze what might contribute to this, we first define 42 features for our route instances, from which we then perform multiple linear regression to see which features would influence the score most. We then choose the set of significant features and execute the Support Vector Machine method to see whether we can have a clear separation between good and bad instances using these chosen features.

\subsection{Feature Selection}

We have created 42 features, all of which are normalized into $[0,1]$. The features can be classified into the following categories:

\textbf{Route features:} There are 5 features defined at the route level. The capacity of a vehicle, the number of zones, stops and packages of a route, and the total travel time of the actual sequence.

\textbf{Zone features:} There are 18 features defined at the zone level. The stop number distribution (mean, standard deviation, maximum and minimum) and package number distribution among zones, the average travel time distribution between two zones (the average travel time from zone 1 to zone 2 is defined as the average travel time from any stops in zone 1 to any stops in zone 2), the average travel time distribution from depot to all zones, and the average travel time from depot to the first visiting zone and the last visiting zone.

\textbf{Stop features:} There are 19 features defined at the stop level. The package volume distribution and package service time distribution of stops, the travel time distribution from depot to all stops, the mean, standard deviation, and maximum package number of stops, the mean, standard deviation, and minimum travel time between two stops, and the travel time from depot to the first visiting stop.

\subsection{Multiple Linear Regression}

We define the logarithm of route score as the dependent variable, and use these 42 features as the independent variables. In total, we have included 2,718 high-quality route instances in our regression. We choose the independent variables that are statistically significant ($p$-value less than 0.01) to be the final set of candidate features. These chosen features are: the number of stops (denoted as \textit{stop\_number}), the total travel time of the actual sequence (denoted as \textit{actual\_seq\_cost}), the average travel time from depot to the first visiting zone (denoted as \textit{depot\_first\_zone}), the average travel time from depot to the last visiting zone (denoted as \textit{depot\_last\_zone}), the mean package volume of stops (denoted as \textit{mean\_pac\_volume}), the standard deviation package volume of stops (denoted as \textit{std\_pac\_volume}), the standard deviation travel time from depot to all stops (denoted as \textit{std\_depot\_stops}), and the standard deviation travel time between two stops (denoted as \textit{std\_tra\_stops}). 

We perform the multiple linear regression again using these 8 features as independent variables, and the result is summarized in Table \ref{tab:regression result}. The features \textit{stop\_number}, \textit{depot\_first\_zone}, \textit{std\_pac\_volume} and \textit{std\_tra\_stops} have positive effects on the score, while the remaining features have negative effects.
\begin{table}[!htb]
\centering
\caption{Multiple linear regression results (standard errors are in parentheses).}
\resizebox{1\columnwidth}{!}{
\begin{tabular}{cc|cc}
\hline
   Feature     & log\_score  & Feature & log\_score \\
\hline
 stop\_number      & \makecell[c]{-1.8365** \\ (0.129)} &  mean\_pac\_volume  & \makecell[c]{2.3787** \\ (0.528)} \\
 actual\_seq\_cost  & \makecell[c]{3.2024** \\ (0.154)}  & std\_pac\_volume  & \makecell[c]{-1.6656** \\ (0.539)} \\
 depot\_first\_zone & \makecell[c]{-1.9557** \\ (0.118)}  & std\_depot\_stops  & \makecell[c]{0.5127* \\ (0.208)} \\
 depot\_last\_zone  & \makecell[c]{1.6719** \\ (0.125)}  & std\_tra\_stops  & \makecell[c]{-2.5941** \\ (0.207)} \\
\hline
 No. of observations  & 2718 \\
\hline
 \multicolumn{2}{l}{\quad$\ast\ast$: p\textless0.01; $\ast$: p\textless 0.05.}
\end{tabular}
}
\label{tab:regression result}
\end{table}

\subsection{Separating Instances using the Support Vector Machine Approach}

Finally, we use the Support Vector Machine (SVM) approach to see whether these 8 features from the route instances can help us distinguish the route instances with good and bad performances. To focus on extreme cases, we define the best-performing instances for routes whose scores are less than 0.01 (240 routes), and the worst-performing instances for routes whose scores are more than 0.1 (199 routes). They are labeled as \emph{low-score} and \emph{high-score} classes respectively. We again divide them into 80\% training set and 20\% testing set to evaluate the feasibility of using SVM as a separation approach.
\begin{table}[!htb]
\centering
\caption{SVM result.}
\begin{tabular}{cccc}
\hline
        & Precision   & Recall & F1-score          \\
\hline
 {lowest scores}   & 0.83      & 0.98    & 0.90          \\
 {highest scores}  & 0.97      & 0.79    & 0.87              \\
 {accuracy}   & /    & /  & 0.89 \\
 {macro avg}   & 0.90    & 0.88  & 0.88 \\
 {weighted avg}   & 0.90    & 0.89  & 0.89 \\
\hline
\end{tabular}
 \label{tab:SVM result}
\end{table}
\begin{table}[!htb]
\centering
\caption{Coefficient of the linear kernel.}
\resizebox{1\columnwidth}{!}{
\begin{tabular}{cc|cc}
\hline
  Feature      & Coefficient   & Feature & Coefficient          \\
\hline
 {stop\_number}   & -3.27  & {mean\_pac\_volume}    & 1.05         \\
 {actual\_seq\_cost}  & 4.80      & {std\_pac\_volume}   & 0.37             \\
 {depot\_first\_zone}   & -3.97    & {std\_depot\_stops}  &  1.78 \\
 {depot\_last\_zone}   & 2.46   & {std\_tra\_stops}  & -1.69 \\
  \hline
\end{tabular}
}
\label{tab:coefficient of linear kernel}
\end{table}

To allow us to interpret the learning outcomes we adopt the \emph{linear kernel} for our SVM process. As we can see from Table \ref{tab:SVM result}, the precision and recall are high, which means the two classes of route instances can be well separated by using these 8 features. The precision of the class `high-score' is 0.97, implying that the poorly-performed route instances can be easily identified. The coefficient of the linear kernel is summarized in Table \ref{tab:coefficient of linear kernel}.

\begin{table*}[!htb]
\centering
\caption{Summary statistics and the mean difference tests.}
  \begin{tabular}{cccccc}
\hline
  feature  & \makecell[c]{lowest scores \\ mean}  &  \makecell[c]{highest scores \\ mean}  & difference & $t$-stat & $p$-value   \\
\hline
 stop\_number        &0.61   & 0.5  & 0.11 &6.12  & \textless 0.01      \\
 actual\_seq\_cost   &0.25   &0.38  &-0.13 &-9.52 & \textless 0.01      \\
 depot\_first\_zone  &0.85   &0.7   &0.15  &12.83 & \textless 0.01      \\
 depot\_last\_zone   &0.76   &0.77  &-0.01 &-0.71 & \textbf{0.48}       \\
 mean\_pac\_volume   &0.08   &0.12  &-0.04 &-6.59 & \textless 0.01      \\
 std\_pac\_volume    &0.06   &0.09  &-0.03 &-5.6  & \textless 0.01      \\
 std\_depot\_stops   &0.16   &0.24  &-0.08 &-7.38 & \textless 0.01      \\
 std\_tra\_stops     &0.3    &0.34  &-0.04 &-4.07 & \textless 0.01      \\
  \hline
\end{tabular}
\label{tab:hypothesis testing}
\end{table*}

\begin{figure*}[!htb]
  \centering
  \includegraphics[width=7in]{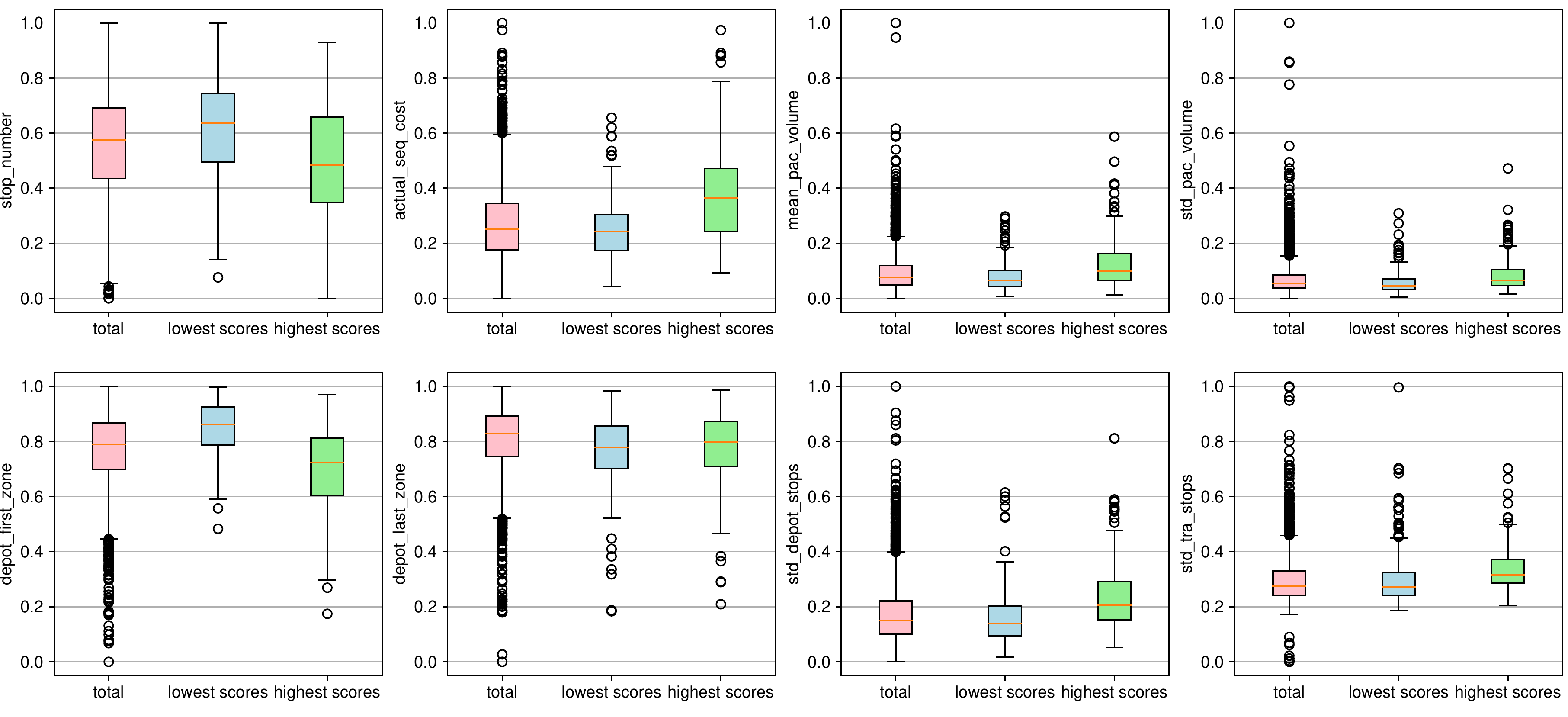}
  \caption{Summary statistics of feature values: total route scores, lowest scores $(<0.01)$, highest scores $(\geq 0.1)$.}
  \label{Box plot}
\end{figure*}

\subsection{Observations}


To see how the mean values of these 8 features are different in the low-score and the high-score classes, we conduct the mean difference test and see that almost all feature means are statistically different, with $p$-values < 0.01 (the only exception is \textit{depot\_last\_zone}, where the $p$-value is 0.48, results summarized in Table \ref{tab:hypothesis testing}). The distributions of feature values are also visualized using box plots in Figure \ref{Box plot}. Based on these analyses, we conclude that low-score instances typically have fewer stops, larger actual total travel time, shorter distance from the depot to the first zone, larger package sizes on average, and greater variations in package sizes, the travel time between stops, and from depot to stops.

In summary, we investigate whether route-level features would have an impact on the performance of our HR-LP approach, and we see from our analyses that the actual total travel time of a given route sequence, the average travel time from the depot to the last visiting zone, the mean package volume of stops, and the standard deviation of travel time from depot to stops have a positive correlation with route scores while the number of stops, the standard deviation of package volumes of stops, the average time from depot to the first visiting zone, and standard deviation of travel time between stops all have a negative correlation with scores. Therefore, given the values of these route features, we can predict whether our approach would perform well.

\section{Conclusion}
In this paper, we propose a novel hierarchical route optimizer with learnable parameters that combines the strength of both the optimization and machine learning approaches. Through numerical evaluations using real-world data, we demonstrate that it is crucial to have optimization and machine learning working together. Besides solving real-world instances well, we also demonstrate how we can use route-related features to identify instances we might have difficulty with. This paves the way to further research on how we can tackle these difficult instances.

\begin{acks}
This research is supported in part by the Ministry of Education, Singapore, under its Social Science Research Thematic Grant (Grant Number MOE2020-SSRTG-018).
\end{acks}



\bibliographystyle{ACM-Reference-Format} 
\bibliography{ref}


\end{document}